\documentclass{article}
\usepackage{spconf,amsmath,graphicx,multirow,booktabs}

\title{Independent language modeling architecture for end-to-end ASR}

\name{%
\parbox{0.9\linewidth}{\centering
      Van Tung Pham$^{1}$,
      Haihua Xu$^1$,
      Yerbolat Khassanov$^{1,2}$,
      Zhiping Zeng$^1$,
      Eng Siong Chng$^{1}$,
      Chongjia Ni$^3$,
      Bin Ma$^3$ and
      Haizhou Li$^4$%
    }%
}
\address{$^1$School of Computer Science and Engineering, Nanyang Technological University, Singapore\\
  $^2$ISSAI, Nazarbayev University, Kazakhstan\\
  $^3$Machine Intelligence Technology, Alibaba Group\\
  $^4$Department of Electrical and Computer Engineering, National University of Singapore, Singapore}
\begin{document}
%
\maketitle
\begin{abstract}

The attention-based end-to-end (E2E) automatic speech recognition (ASR) architecture allows for joint optimization of acoustic and language models within a single network.
However, in a vanilla E2E ASR architecture, the decoder sub-network (subnet), which incorporates the role of the language model (LM), is conditioned on the encoder output.
This means that the acoustic encoder and the language model are entangled that doesn't allow language model to be trained separately from external text data. To address this problem, in this work, we propose a new architecture that separates the decoder subnet from the encoder output. In this way, the decoupled subnet becomes an independently trainable LM subnet, which can easily be updated using the external text data. We study two strategies for updating the new architecture. 
Experimental results show that, 1) the independent LM architecture benefits from external text data, achieving 9.3\% and 22.8\% relative character and word error rate reduction on Mandarin HKUST and English NSC datasets respectively; 2) the proposed architecture works well with external LM and can be generalized to different amount of labelled data.

\end{abstract}
\section{Introduction}
\label{sec:intro}

End-to-End (E2E) architecture has been a promising strategy for ASR systems. In this strategy, a single network is employed to directly map acoustic features into a sequence of characters or words without the need of a pronunciation dictionary that is required by the conventional Hidden Markov Model based systems. Furthermore, the components of E2E network can be jointly trained for a common objective criterion to achieve overall optimization. 
The main approaches for E2E ASR are attention-based encoder-decoder  \cite{ChanJLV16,DBLP:conf/icassp/BahdanauCSBB16,46687,NIPS2015_5847,DBLP:conf/interspeech/ChorowskiJ17,46169}, Connectionist Temporal Classification (CTC) \cite{Graves:2014:TES:3044805.3045089,pmlr-v48-amodei16} and the hybrid CTC/attention architectures \cite{kim2017joint,watanabe2017hybrid}.

The training of an E2E system requires a large amount of transcribed speech data, here denoted as labelled data, which is unavailable for low-resource languages. However, we note that large external text data can easily be collected. In this work, we focus on the use of external text data to improve the language model (LM) of E2E ASR systems. 

In a vanilla E2E architecture, the decoder sub-network (subnet) incorporates the role of the LM. Unlike traditional ASR systems where the LM is separated and hence can easily be trained with text-only data, the decoder subnet is conditioned on the encoder output. As a result, it is not straightforward to update the LM component of the vanilla E2E architecture with the text data. 

To address this problem, in this work, we introduce a new architecture which separates the decoder subnet from the encoder output, making the subnet an explicit
LM. In this way, the subnet can easily be updated using the text data. A potential issue, named catastrophic forgetting \cite{catastrophicForget}, might occur when using external text to update the E2E network: the network forgets what it has learnt from labelled data. We, therefore, study the strategies that use both labelled and external text data to update the E2E network\footnote{Since both labelled and external text data are used, we actually allow entire E2E network to be updated.}. 

The paper is organized as follows. Section \ref{trainE2E} describes a vanilla architecture of E2E ASR systems. In Section \ref{proposedArchitecture}, we first describe the proposed architecture
, then present strategies to update the proposed architecture using external text data. Section \ref{RelatedWork} relates our proposed approach with prior work. Experimental setup and results are presented in Section \ref{expSetup} and \ref{expResults} respectively. Section \ref{conclusions} concludes our work.

\section{A vanilla E2E ASR architecture}
\label{trainE2E}
In this section, we describe a vanilla attention-based\footnote{In actual implementation, we use the hybrid CTC/attention architecture \cite{kim2017joint,watanabe2017hybrid}. However, since the CTC module is untouched, we do not mention it during the rest of this paper for simplicity.} E2E ASR architecture (denoted as $A1$), which is widely used in prior work \cite{kim2017joint,watanabe2017hybrid}. 
Let first denote $\mathcal{P}$ as the labelled data. Let $<$\textbf{X}, \textbf{Y}$>$ $\in$ $\mathcal{P}$ be a training utterance, where $\textbf{X}$ is a sequence of acoustic features and $\textbf{Y} = \{y_1, y_2,..., y_{|\textbf{Y}|} \}$ is a sequence of output units. 

\begin{figure}[h]
  \centering
\includegraphics[scale=0.42,trim = 27mm 97mm 23mm 70mm,clip]{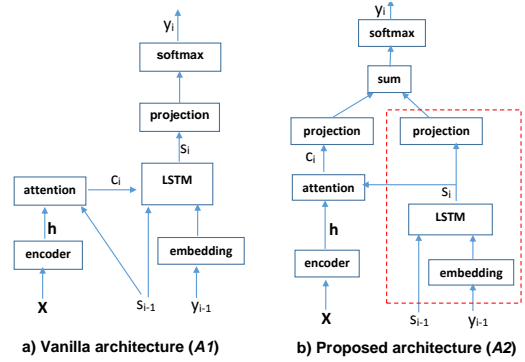}
  \caption{(a) A vanilla E2E architecture $A1$ and (b) the proposed architecture $A2$. In the proposed architecture, the subnet (in red-dash box) is a language model which can be easily updated using external text data.}
  \label{s2cModel}
\end{figure}

The E2E architecture consists of an encoder and an attention-based decoder which are shown in Fig. \ref{s2cModel} (a). The encoder acts as an acoustic model which maps acoustic features into an intermediate representation \textbf{h}. Then, the decoder subnet, which consists of an embedding, a Long Short-Term Memory (LSTM) and a projection layers, generates one output unit at each decoding step $i$ as follows,
\begin{align}
    c_i &= attention(\textbf{h}, s_{i-1}) \label{conVec} \\
    s_i &= LSTM(s_{i-1}, c_i, embedding(y_{i-1})) \label{decoderHidden} \\
    P(y_i \mid \textbf{X}, y_{<i}) &= softmax(projection(s_i)) \label{softmax}
\end{align}
where $c_i$ is the context vector, $s_{i-1}$ and $s_{i}$ are output hidden states at time step $i-1$ and $i$ respectively, $embedding()$ and $projection()$ are embedding and projection layers respectively. The E2E network is normally trained in batch-mode with a loss function as follows,
\begin{equation}\label{loss}
\begin{split}
    L_{ASR}(\theta) & = \frac{1}{|B|} \sum\nolimits_{<\textbf{X},\textbf{Y}> \in B} \log P(\textbf{Y} \mid \textbf{X}, \theta) \\
    & = \frac{1}{|B|}\sum\nolimits_{<\textbf{X},\textbf{Y}> \in B} \sum_{i=1}^{|\textbf{Y}|} \log P(y_i \mid \textbf{X}, y_{<i}, \theta)
\end{split}
\end{equation}
where $y_{<i}$, $B$ and $\theta$ denote the decoding output history, a batch of data and model parameters respectively.

According to Equation \eqref{conVec} and \eqref{decoderHidden}, the LSTM is conditioned on the context vector $c_i$ which depends on the encoder output $\textbf{h}$. In the absence of acoustic features $\textbf{X}$, thus $\textbf{h}$, it is not possible to update the E2E architecture appropriately using only text data. One way to alleviate such problem is to set $c_i$ by an all-zero vector \cite{DBLP:journals/corr/abs-1808-10128}. Unfortunately, this method introduces a mismatch between training phase and updating phase (with external text data) since during training $c_i$ is generally not the all-zero vector.
\section{Independently trainable LM subnet}
\label{proposedArchitecture}
To allow updating of LM with external text data, we first introduce (Section \ref{newArchitecture}) a novel architecture that separates the decoder subnet from the encoder output. The updating algorithm is described in Section \ref{updateDecoder}.
\subsection{Decoupling LM subnet}
\label{newArchitecture}
Inspired by the idea of spatial attention \cite{8099828} for image captioning, we propose to decouple the LM subnet from the encoder output as shown in Fig. 1(b). In this architecture, denoted as $A2$, the decoding process is formally described as follows,
\begin{align}
    s_i &= LSTM(s_{i-1}, embedding(y_{i-1})) \label{decoderNew}\\
    c_i &= attention(\textbf{h}, s_{i}) \label{attentionNew}\\
        P(y_i \mid \textbf{X}, y_{<i}) =\ & softmax(projection(s_i)\ + \nonumber \\
                                     & projection(c_i)) \label{distrNew}
\end{align}

From Equation \eqref{decoderNew}, the LSTM is only conditioned on the previous decoding hidden state and previous decoding output. In other words, the decoder subnet is a standard LM, hereafter denoted as LM subnet. In this way, this subnet can be independently updated with external text data. 
\subsection{Updating the LM subnet with external text data}
\label{updateDecoder}
\begin{figure}[h]
  \centering
  \includegraphics[scale=0.5,trim = 45mm 170mm 70mm 45mm]{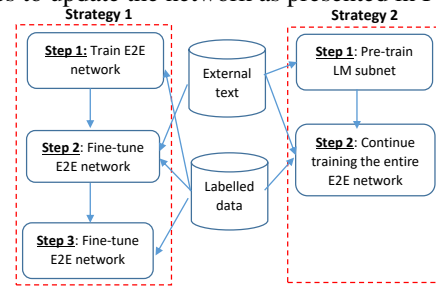}
  \caption{Strategies to update the E2E architecture with external text data.}
  \vspace{-0.2cm}
  \label{fine_tune}
\end{figure}

One issue when using external text, denoted as $\mathcal{T}$, to improve LM is catastrophic forgetting. Specifically, when $\mathcal{T}$ is used, the network forgets what it has learnt from $\mathcal{P}$. To address such issue, we use both $\mathcal{T}$ and $\mathcal{P}$ to update the entire E2E network. 
Another issue is when should $\mathcal{T}$ be used, i.e. before or after the entire E2E ASR network is trained. We study two strategies to update the network as presented in Fig. \ref{fine_tune}. 

In Strategy 1, the entire E2E network is first trained using $\mathcal{P}$. Then, in the second step, the network is fine-tuned with both $\mathcal{T}$ and $\mathcal{P}$. Finally, the network is further fine-tuned with the data $\mathcal{P}$. We empirically found that the last step improves the system performance. 

In Strategy 2, the LM subnet is pre-trained with the text data $\mathcal{T}$ first. Then, the entire E2E network is trained using both $\mathcal{T}$ and $\mathcal{P}$.  

In the second step of both strategies, i.e. to use both $\mathcal{T}$ and $\mathcal{P}$ to update the E2E network, the following loss function is used:
\begin{equation}\label{totalLoss}
    L_{total}(\theta) = (1 - \alpha) L_{ASR}(\theta)  +  \alpha L_{LM}(\theta_{d})
\end{equation}
where $\alpha$ denotes an interpolation factor, $\theta_{d}$ denotes all LM subnet parameters, $L_{LM}(\theta_{d})$ denotes the LM loss obtained when external text data is used, i.e.: 
\begin{equation}\label{loss}
\begin{split}
    L_{LM}(\theta_{d}) & = \frac{1}{|B_1|} \sum\nolimits_{\textbf{Y'} \in B_1} \log P(\textbf{Y'} \mid \theta_{d}) \\
    & = \frac{1}{|B_1|}\sum\nolimits_{\textbf{Y'} \in B_1} \sum_{i=1}^{|\textbf{Y'}|} \log P(y_i \mid y_{<i}, \theta_{d})
\end{split}
\end{equation}
where $B_1$ is a batch of external text data.
\vspace{-0.2cm}
\section{Comparison with Related work}
\label{RelatedWork}
There have been several studies on how to use external text data for E2E ASR. One of the ideas is to use the external text data to build an external LM, then incorporate it into the inference process \cite{Graves:2014:TES:3044805.3045089,Hori2018} or employ it to re-score n-best output hypotheses \cite{ChanJLV16}. Our proposed approach, however, improves the language modeling capability of an E2E system without using any external LM. Another idea is data synthesis. Specifically, \cite{JinxICASSP2019} used a text-to-speech system while \cite{RenduchintalaInterspeech2018} used a pronunciation dictionary and duration information to generate additional inputs given the external text, which are then used to train a correction model \cite{JinxICASSP2019} or another encoder \cite{RenduchintalaInterspeech2018}. Our proposed approach uses external text data without involving external systems, such as text-to-speech.


In natural language processing, exploiting text corpora to improve E2E systems is also widely used. A popular approach is to use a text corpus to pre-train entire E2E network \cite{NIPS2015_5949,SongTQLL19}. Such techniques are only applicable for tasks where both input and output are in text format. Another approach is to pre-train only the decoder by simply removing the encoder \cite{NIPS2015_5949,45824}. This is equivalent to zeroing out the context vector \cite{DBLP:journals/corr/abs-1808-10128},
which introduces a mismatch as discussed in Section \ref{trainE2E}.

The idea of separating the decoder from encoder output has been introduced in image captioning research community \cite{8099828}. 
To the best of our knowledge, this work is the first attempt applying it on the ASR task.
\vspace{-0.2cm}
\section{Experimental setup}
\label{expSetup}
\subsection{Corpora}
We conduct experiments on LDC2005S15, which is the HKUST Mandarin Telephone Speech \cite{conf/iscslp/LiuFYCHG06}, and the National Speech Corpus (NSC) \cite{koh2019building}, which is a Singapore English microphone data set.

The HKUST corpus consists of 171.1 hours of Mandarin Chinese conversational telephone speech from Mandarin speakers in mainland China. It is divided into a training set of 166.3 hours, and a test set of 4.8 hours. We split the training data into 3 subsets: the first two subsets are $\mathcal{P}$ and $\mathcal{T}$ in this paper, while the remaining subset is used for validation. The detailed information of these data sets is presented in Table \ref{datasets}. For the labelled data $\mathcal{P}$, we perform speed-perturbation based data augmentation \cite{DBLP:conf/interspeech/KoPPK15}.  
We report Mandarin character error rate (CER) on the test set. 

The NSC corpus consists of 2,172.6 hours of English read microphone speech from 1,382 Singaporean speakers. We extract data of 6 speakers as testing data. Similar to the HKUST corpus, we split the remaining data into 3 subsets for $\mathcal{P}$, $\mathcal{T}$ and validation. The detailed data division is shown in Table \ref{datasets}. We also perform data augmentation on the labelled data $\mathcal{P}$. We report word error rate (WER) on the test set.
\vspace{-0.2cm}
\begin{table}[]
\small
\centering
\begin{tabular}{l|c|c|c|c}
\toprule
\multirow{3}{*}{Datasets}           & \multicolumn{2}{c|}{HKUST}                                           & \multicolumn{2}{c}{IMDA}                                               \\ \cline{2-5} 
                                    & \#utts  & \begin{tabular}[c]{@{}l@{}}Duration\\ (hours)\end{tabular} & \#utts  & \begin{tabular}[c]{@{}l@{}}Duration\\ (hours)\end{tabular}   \\ \hline
Labelled data ($\mathcal{P}$)       & 22,500  & 20.2                                                       & 15,000  & 20.6                                                         \\ \hline
External text ($\mathcal{T}$)       & 158,605 & -                                                          & 1,547,399 & -                                                          \\ \hline
Validation                          & 5,457   & 4.88                                                       & 16,144   & 21.2                                                        \\ \hline
Test                                & 5,151   & 4.75                                                       & 5,589   & 7.6                                                          \\
\bottomrule
\end{tabular}
\caption{Data division for the HKUST and NSC corpora.}
\label{datasets}
\vspace{-0.2cm}
\end{table}

\subsection{E2E configuration}
\label{e2eConfig}
We use the ESPnet toolkit \cite{DBLP:conf/interspeech/WatanabeHKHNUSH18} to develop our E2E models. We use 80 mel-scale filterbank coefficients with pitch as  input features. The encoder consists of 6 layers VGG \cite{Hori2017AdvancesIJ} and 6 layers BLSTM, each has 320 units. In this paper, we used the location-aware attention mechanism \cite{NIPS2015_5847}. Characters and Byte-Pair Encoding (BPE) (500 units) are used as output units for HKUST and NSC corpora respectively.

We set the batch size for training data $\mathcal{P}$ as $B = 30$.  Since $\mathcal{T}$ has many more utterances than $\mathcal{P}$, we set the batch size for text data as  $B_1$ = 150 and 300  for HKUST and NSC respectively. The optimizer is the AdaDelta algorithm \cite{Zeiler2012ADADELTAAA} with gradient clipping \cite{Pascanu2012UnderstandingTE}. We used $\lambda = 0.1$ for both corpora. During decoding, we used beam width 30 for all conditions.

\section{Experimental results}
\label{expResults}
\subsection{Independent LM architecture and updating strategies}
\begin{figure}[!htb]

\begin{minipage}[b]{1.0\linewidth}
  \centering
  \centerline{\includegraphics[scale=0.6,trim = 0mm 0mm 0mm 0mm,clip]{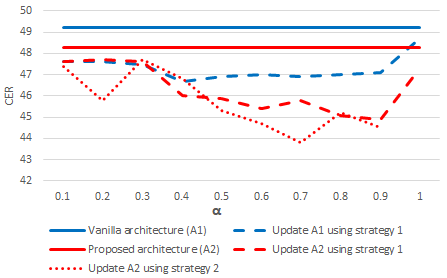}}
  \centerline{(a) Results on the HKUST corpus}\medskip
\end{minipage}
\begin{minipage}[b]{1.0\linewidth}
  \centering
  \centerline{\includegraphics[scale=0.6,trim = 0mm 0mm 0mm 0mm,clip]{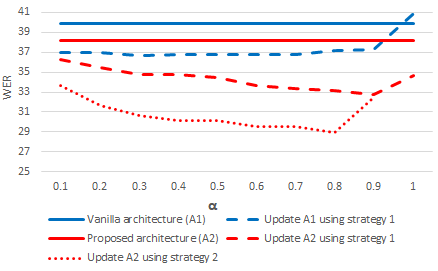}}

  \centerline{(b) Results on the NSC corpus}\medskip
\end{minipage}
\caption{Comparison between the vanilla architecture $A1$ (blue color), and the proposed architecture $A2$ (red color). Solid lines denote the two architectures trained using only $\mathcal{P}$ while dashed and dotted lines indicate these architectures updated with external text data $\mathcal{T}$ at different values of the factor $\alpha$ (see Eq. \eqref{totalLoss}) using two strategies described in Section \ref{updateDecoder}.}
\label{compareS1S2}
\end{figure}

In this section, we compare the vanilla architecture $A1$ to the proposed architecture $A2$ when they are trained using labelled data $\mathcal{P}$. We then compare two updating strategies described in Section \ref{updateDecoder} when external text data $\mathcal{T}$ is used. To update $A1$ with the external text data, we set the context vector by an all-zero vector \cite{DBLP:journals/corr/abs-1808-10128} as mentioned in Section \ref{trainE2E}. The results are presented in Fig. \ref{compareS1S2}. We have following observations. 
\begin{itemize}
  \item The proposed $A2$  consistently outperforms $A1$ on both HKUST and NSC corpora. Particularly, $A2$ outperforms $A1$ by 1.8\% relative (from 49.2\% to 48.3\%) CER and 4.3\% relative (from 39.9\% to 38.2\%) WER on HKUST and NSC corpora respectively. 
  \item With external text data, Strategy 1 leads to significant error rate reduction for both architectures. For example, on the $A2$ architecture, at $\alpha = 0.9$ we observe 14.4\% relative (from 38.2\% to 32.7\%) WER 
  reduction for the NSC corpus. We also observe that $A2$ outperforms $A1$ for all cases which indicates that $A2$ benefits more from the external text.
  \item Strategy 2 generally outperforms Strategy 1 when they are applied on $A2$. At $\alpha=0.7$, Strategy 2 (denoted as $A2$-$Strategy2$-$0.7$) achieves the best results on two corpora, i.e. 43.8\% CER and 29.5\% WER (which are 9.3\% relative CER and 22.8\% relative WER reduction over $A2$) on HKUST and NSC respectively. We will use $A2$-$Strategy2$-$0.7$ for experiments in the next section.
\end{itemize}

\vspace{-0.2cm}
\subsection{Interaction with external LM}
\begin{figure}[!h]
  \centering
  \includegraphics[scale=0.7]{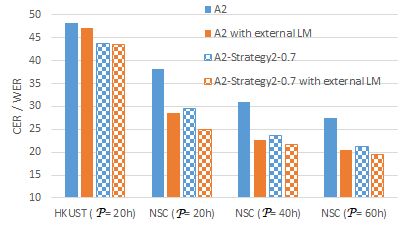}
  \caption{ASR performance with and without external LM at different amount of labelled data. }
  \vspace{-0.4cm}
  \label{analysis}
\end{figure}

In this section, we first show the interaction between the proposed independent LM architecture and external LM \cite{Graves:2014:TES:3044805.3045089,Hori2018}. Specifically, we train a Recurrent Neural Network LM (RNN-LM) as a 1-layer LSTM with 1000 cells for both corpora, then integrate the RNN-LM into inference process of $A2$ and $A2$-$Strategy2$-$0.7$. We also examine the effect of varying amount of labelled data $\mathcal{P}$ from 20 hours to 60 hours. For this experiment, we only conduct on NSC corpus since the HKUST is relatively small. Results are reported in Fig \ref{analysis}. 

We observe that the external RNN-LM improves $A2$-$Strategy2$-$0.7$ by 0.2\% absolute CER and 4.5\% absolute WER on 20 hours of HKUST and NSC corpora respectively. The results indicate that our proposed approach benefits from the external LM. Additionally, we observe consistent improvements at different amount of $\mathcal{P}$ on NSC, which demonstrate that our proposed architecture works well under different amount of labelled data.
\vspace{-0.2cm}
\section{Conclusions}
\label{conclusions}
We introduced a new architecture that separates the decoder subnet from the encoder output so that it can be easily updated using an external text data. Experimental results showed that the new architecture not only outperforms the vanilla architecture when only labelled data is used, but also benefits from the external text data. We studied two strategies to update the E2E network and found that by pre-training the subnet with the text data then fine-tuning the entire E2E network using both labelled and text data, we achieve the best results. Further analyses also showed that the proposed architecture can be augmented with an external LM for further improvement and can be generalized with different amount of labelled data. 
\vspace{-0.5cm}
\section{Acknowledgements}
This work is supported by the project of Alibaba-NTU Singapore Joint Research Institute.



\bibliographystyle{IEEEbib}
\bibliography{strings,refs}

\end{document}